\documentclass[letterpaper,10pt,conference]{ieeeconf}
\IEEEoverridecommandlockouts
\usepackage[T1]{fontenc}
\usepackage[utf8]{inputenc}
\usepackage{amsmath,amssymb}
\usepackage{graphicx}
\graphicspath{{figures/}}
\usepackage{array,booktabs}
\usepackage{url}
\usepackage{cite}
\usepackage{xcolor}
\makeatletter
\let\NAT@parse\undefined
\makeatother
\usepackage[colorlinks=true,citecolor=blue,linkcolor=blue,urlcolor=black]{hyperref}
\newcolumntype{L}[1]{>{\raggedright\arraybackslash}p{#1}}

\title{\LARGE\bf MagPillar: A Soft Magnetopillar Array Tactile Sensor for Robotic Surface Perception}
\author{Haotian Guo$^{\dagger}$, Yu Feng$^{\dagger}$, Hao Wu$^{\dagger}$, Jiazhi Tang, Ningbin Zhang, and Jianshu Zhou$^{*}$
\thanks{$^{\dagger}$Haotian Guo, Yu Feng, and Hao Wu contributed
equally to this work.}%
\thanks{$^{*}$Corresponding author: Jianshu Zhou (e-mail: jianshuzhou@nus.edu.sg).}%
\thanks{Haotian Guo, Yu Feng, Hao Wu, Jiazhi Tang, and Jianshu Zhou
are with the Department of Mechanical Engineering,
National University of Singapore, Singapore. Ningbin Zhang is with School of Mechanical Engineering, Shanghai Jiao Tong University, Shanghai 200240, China}%
}
\begin{document}
\bstctlcite{MagPillarBibControl}
\maketitle
\thispagestyle{empty}
\pagestyle{empty}
\raggedbottom

\begin{abstract}
Robotic surface exploration benefits from tactile interfaces that accommodate contact and capture variations in surface characteristics. Herein, we report MagPillar, a tactile sensor inspired by spider tactile hairs that combines a magnetic pillar array with three-axis Hall sensing. A full-array mechanical and magnetic-field model characterizes how pillar deformation affects the superposed fields measured across the array, providing a physical basis for interpreting its contact responses. The compliant pillar architecture translates contact with surface textures and geometric features into distinct magnetic response patterns. Then, in robotic surface perception experiments, spectral features and logistic regression distinguish six material surfaces with an offline accuracy of 99.34\% under within-session temporally blocked cross-validation. Continuous scans over periodic protrusions produce amplitude and peak-spacing variations consistent with changes in surface geometry. These findings connect compliant contact mechanics with distributed magnetic sensing, paving a new path for material recognition and qualitative surface-profile perception.
\end{abstract}


\section{Introduction}
\label{sec:introduction}

\PARstart{T}{actile} sensing provides robots with information about object surfaces through direct physical interaction~\cite{zhou2026everything,zhou2024dexterous}. During sliding, surface roughness, compliance, and local geometry affect contact forces and deformation, producing signals that can be used to distinguish materials and identify surface features. For example, biomimetic fingertips combine vibration and thermal measurements for material classification~\cite{Kerr2018Material}, while optical tactile sensors such as GelSight image the deformation of a compliant contact surface to resolve local geometry and texture~\cite{Yuan2017GelSight}. For robotic surface exploration, the contact interface and sensing mechanism must therefore be considered together to capture useful surface information.

\begin{figure}[t]
\centering
\includegraphics[width=\columnwidth]{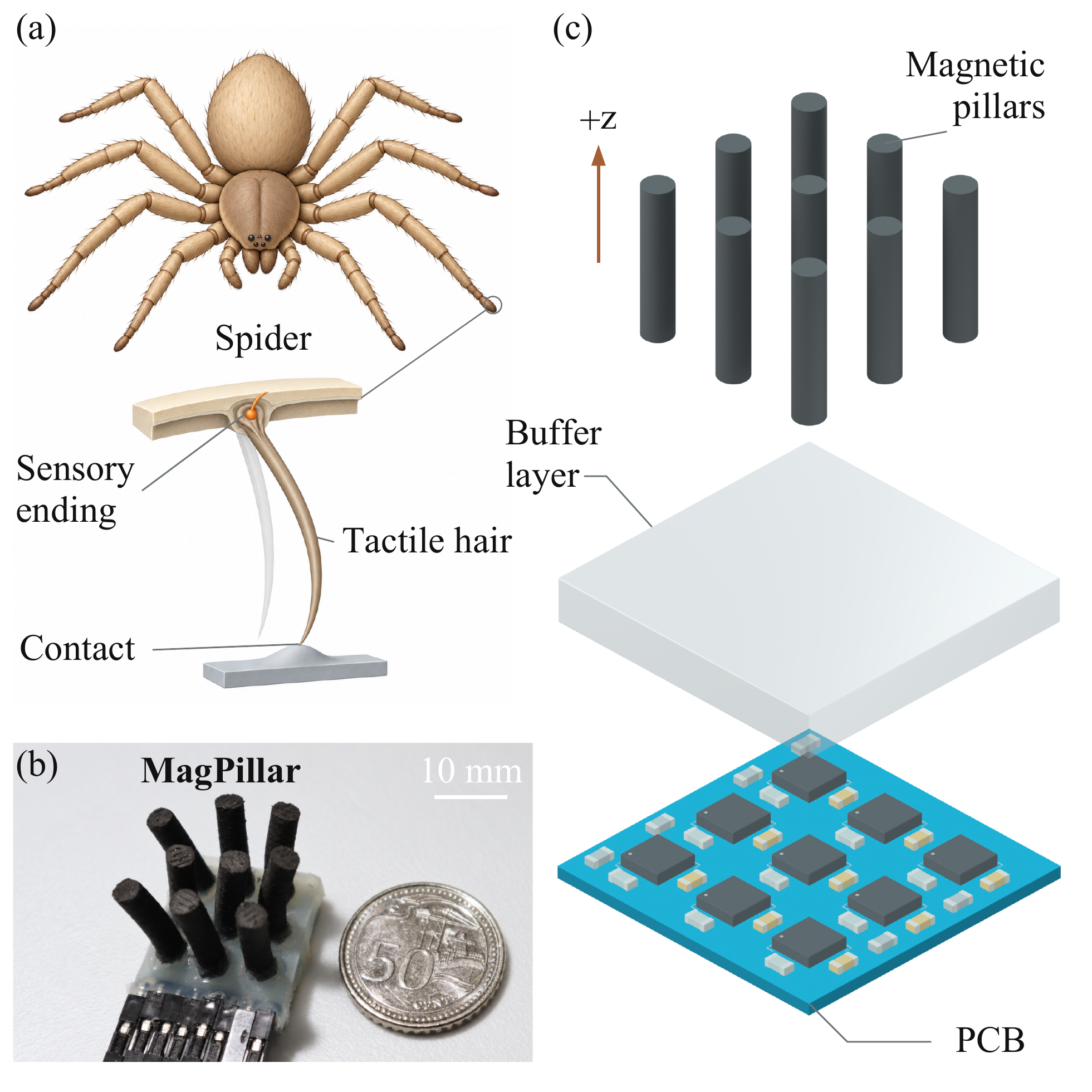}
\caption{Bioinspired design and structure of MagPillar.
(a) Bioinspiration from spider tactile hairs.
(b) Prototype of the $3\times3$ magnetic pillar array.
(c) Exploded view of the magnetic pillars, buffer layer, and PCB.}
\label{fig:prototype}
\end{figure}

Different transduction mechanisms have been used to acquire material and texture information~\cite{wang2026relaxation,feng2026soft,Feng2026RoboTac}. FlexiRay combines a compliant gripper with multimodal visual--tactile sensing to identify surface textures and classify objects~\cite{wang2025flexible}. Fingerprint-inspired piezoelectric sensors encode sliding-induced vibrations as texture-dependent voltage waveforms~\cite{Qin2024Fingerprint}. Iontronic sensors exploit changes in electric-double-layer capacitance to capture both static pressure and high-frequency vibrations, enabling fine textile features to be resolved during sliding~\cite{Bai2023Robotic}. Multimodal electronic skins further combine piezoelectric, triboelectric, and piezoresistive mechanisms to distinguish pressure, strain, and vibration, providing complementary information about material properties and contact interactions~\cite{Lee2021Fingerpad}. These approaches impose different integration requirements: vision-based designs need space for imaging and illumination, whereas electrical skins incorporate sensing materials and electrodes into the deformable interface.

Magnetic tactile sensing converts contact-induced deformation into magnetic-field changes, allowing the deformable interface to remain separate from the readout electronics without embedded wiring or an internal imaging path. For example, MagOne uses a Hall sensor to detect the displacement of a magnet embedded in a soft elastomer and reconstruct three-axis contact forces~\cite{wang2016design}. In magnetic-film designs, sinusoidal magnetization enables decoupled measurement of normal force and one tangential force component~\cite{Yan2021}, while film arrays extend sensing to multiple contact locations~\cite{Hu2024}. Beyond force and contact measurement, magnetic signals recorded during sliding also capture surface-dependent variations. These signals have been used for fabric texture recognition through sequence learning~\cite{Yan2022} and material classification based on time- and frequency-domain features~\cite{Dai2022}.

Protruding magnetic cilia provide another contact structure: sliding interactions bend the cilia and change the magnetic fields measured beneath them. Double-layer magnetic cilia have been used to characterize sandpaper roughness~\cite{Man2023}, while magnetic hair arrays detect sliding and recognize objects from spatial and temporal responses~\cite{Man2024}. MagCilia combines a compliant magnetic structure with a single Hall sensing unit and uses spectral features of reconstructed forces for surface recognition~\cite{Feng2026MagCilia}. These studies motivate the use of deformable protruding elements for surface exploration.

Interpreting their outputs also requires understanding the relationship between deformation and magnetic readout. A two-segment beam model relates cilium loading to magnetic-field changes~\cite{Liu2026}, while a coupled deformation, magnetic-field, and tunnel magnetoresistive model supports bending-angle and direction measurements~\cite{Ribeiro2026}. For a multipillar array, a further question is how local deformation contributes to signals at different sensing locations. Each magnetic sensor receives superposed fields from multiple pillars, so a one-to-one correspondence between a contact element and a readout channel cannot be assumed. An array-level analysis is therefore needed to relate local contact mechanics to the distributed signals used for surface perception.

In this study, we present MagPillar, a passive $3\times3$ compliant magnetic pillar array coupled with nine three-axis Hall sensors for robotic surface exploration. A full-array mechanical and magnetic-field model characterizes how single- and two-pillar loading produces distributed responses across all 27 magnetic-field channels, including a peak response adjacent to the loaded pillar. Robotic scanning experiments use spectral features and logistic regression to distinguish six material surfaces and record amplitude and peak-spacing variations over periodic protrusions with different heights and pitches. Together, the modeling and experiments connect compliant pillar deformation with distributed magnetic readout for material recognition and continuous surface sensing.

\section{Sensor Structure and Fabrication}
\label{sec:fabrication}

\subsection{Bioinspired Sensor Design}
\label{sec:architecture}

\begin{figure}[t]
\centering
\includegraphics[width=0.98\columnwidth]{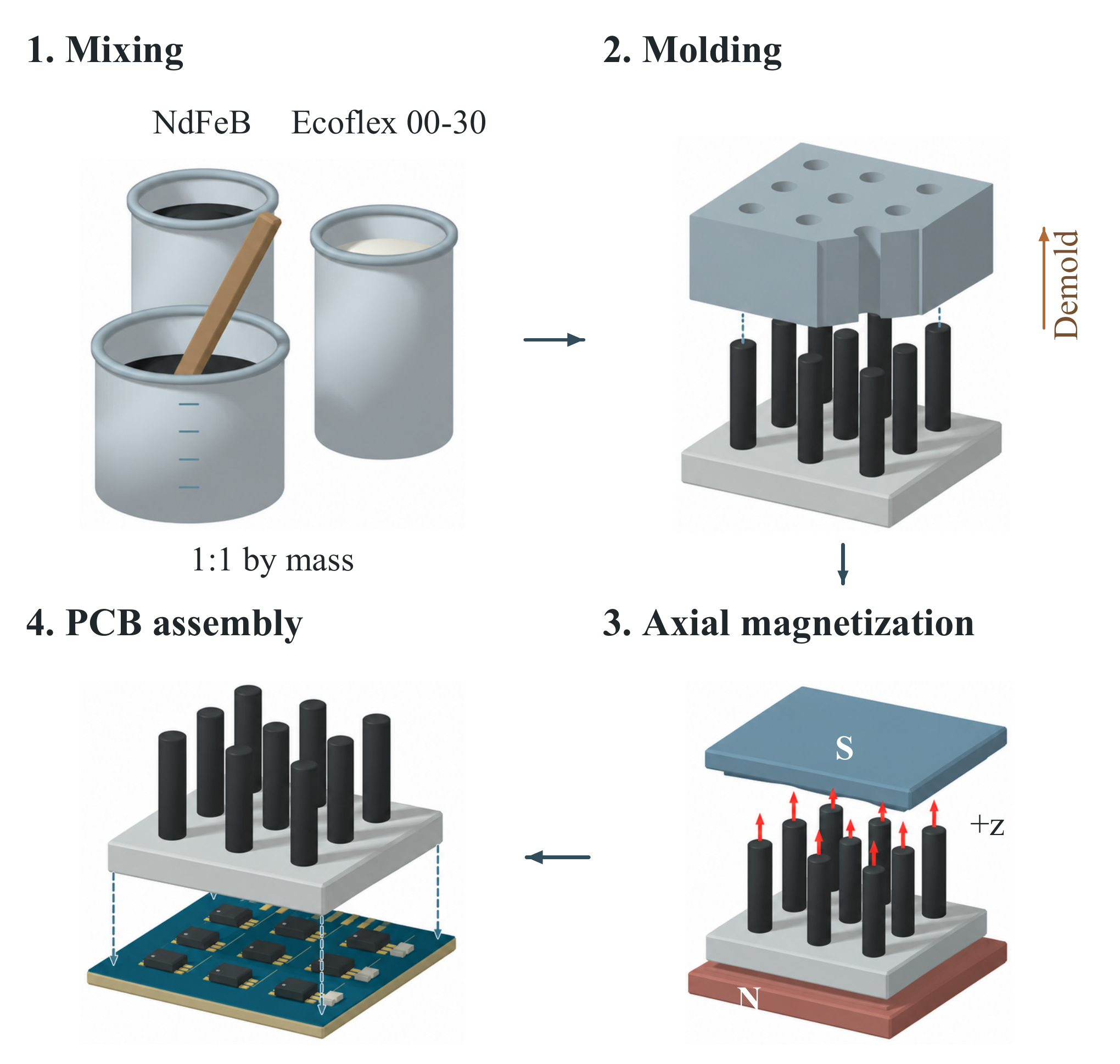}
\caption{Schematic fabrication process of MagPillar. (1) Mixing NdFeB powder and Ecoflex 00-30. (2) Molding and demolding. (3) Axial magnetization along $+z$. (4) Assembly with the Hall sensor PCB.}
\label{fig:fabrication}
\end{figure}

Spider tactile hairs deflect under direct contact and transmit mechanical inputs through their basal structures to sensory terminals~\cite{Barth2004} (Fig.~\ref{fig:prototype}(a)). Inspired by this mechanism, MagPillar combines compliant magnetic pillars with an underlying Hall sensor array to convert contact-induced deformation into measurable magnetic-field changes. The protruding pillars form a deformable contact interface, allowing interactions with surface features to be captured during robotic exploration.

As shown in Figs.~\ref{fig:prototype}(b) and~\ref{fig:prototype}(c), the sensor consists of a magnetic pillar array, a compliant buffer layer, and a printed circuit board (PCB). Nine cylindrical pillars are arranged in a $3\times3$ array, each with a length of 10~mm and a diameter of 2~mm, at a center-to-center pitch of 6~mm. The shared buffer layer supports the pillar roots and provides a compliant interface above the rigid PCB. The PCB carries nine MLX90393 three-axis Hall sensors aligned beneath the pillar centers, with a nominal root-to-Hall separation of 3~mm. The principal geometric parameters are summarized in Table~\ref{tab:device}.

The undeformed pillar axis defines the $z$ direction, while the $x$ and $y$ axes lie in the array plane. All pillars are magnetized axially along $+z$. Contact-induced bending and compression change the positions and orientations of the magnetized material relative to the fixed Hall sensors, thereby modifying the detected magnetic fields. Each Hall sensor measures the components $B_x$, $B_y$, and $B_z$, providing a total of 27 magnetic-field outputs.

The array provides multiple compliant contact sites for sampling surface interactions during scanning. Because each Hall sensor receives magnetic contributions from multiple pillars, local deformation produces a spatially distributed response. The combined three-axis readout captures contact variations across the array, supporting material recognition and surface-profile perception with the electronics beneath the passive contact interface.

\begin{table}[t]
\caption{Geometric parameters of MagPillar}
\label{tab:device}
\centering\small
\begin{tabular*}{\columnwidth}{@{\extracolsep{\fill}}lr@{}}
\toprule
Parameter & Value (mm) \\
\midrule
Pillar length & 10 \\
Pillar diameter & 2 \\
Pillar pitch & 6 \\
Base dimensions & $19.97\times19.97\times3$ \\
Root-to-Hall separation & 3 \\
\bottomrule
\end{tabular*}
\end{table}

\subsection{Materials and Fabrication}
\label{sec:materials_fabrication}

The fabrication process is illustrated in Fig.~\ref{fig:fabrication}. A magnetizable elastomer composite is prepared by mixing 400-mesh neodymium--iron--boron (NdFeB) powder with Ecoflex 00-30 at a powder-to-total-silicone mass ratio of $1:1$. The NdFeB particles provide the magnetizable phase, while the silicone matrix supplies the mechanical compliance required for pillar deformation. The composite is molded into the $3\times3$ pillar array, and the buffer layer is fabricated from unfilled Ecoflex to provide a nonmagnetic, compliant support.

After demolding, the pillar array is magnetized axially along $+z$, establishing a common initial magnetization direction for deformation sensing. The magnetic array and buffer layer are assembled with the Hall sensor PCB, with the pillar centers aligned above the corresponding Hall sensing locations. The resulting structure integrates deformable magnetic contact elements with fixed three-axis magnetic readout.

\section{Mechanics and Magnetic-Field Model}
\label{sec:model}

\subsection{Full-Array Forward Modeling}
\label{sec:forward}

Quasistatic finite element analysis in COMSOL is combined with magnetic-field calculations to characterize the deformation and magnetic response of the $3\times3$ pillar array under lateral and normal loading. The model uses the geometry in Table~\ref{tab:device}, with all nine pillars initially magnetized along $+z$. Nine fixed Hall sampling points are located beneath the pillar centers, 3~mm below the undeformed pillar-root plane.

The pillars and base are modeled as nearly incompressible Neo-Hookean materials with geometric nonlinearity. Their Young's moduli are set to 0.3 and 0.05~MPa, respectively, with a Poisson's ratio of 0.49. The bottom surface of the base is fixed, and forces are applied at selected pillar tips. The reference remanence magnitude is set to $B_r=0.1$~T, and the relative permeability is unity. The remanence vector is updated according to $\mathbf B_r=\mathbf F\mathbf B_{r,0}/J$~\cite{Zhao2019}, where $\mathbf B_{r,0}$ is the initial remanence vector, $\mathbf F$ is the deformation gradient, and $J=\det\mathbf F$. The magnetic field at each Hall sampling point is then calculated using a free-space end-face source integral over the deformed geometry, including contributions from all nine pillars.

The three-axis magnetic-field increments at the nine sampling points are normalized by $B_r$ and assembled into a 27-dimensional response vector, $\Delta\widetilde{\mathbf b}$. Near the unloaded state, this vector is approximated as a linear function of the applied forces:
\begin{equation}
\Delta\widetilde{\mathbf b}
=\frac{1}{B_r}
\left[\Delta\mathbf B_1^{\mathsf T},\ldots,\Delta\mathbf B_9^{\mathsf T}\right]^{\mathsf T}
\simeq\mathbf K\mathbf f,
\label{eq:1}
\end{equation}
where $\Delta\mathbf B_i=\mathbf B_i-\mathbf B_{i,0}$ denotes the field change from the unloaded reference at sampling point $i$. The vector $\mathbf f$ contains the 27 force components applied to the nine pillars, and $\mathbf K\in\mathbb R^{27\times27}$ is the local response matrix with units of $\mathrm{N}^{-1}$.

\subsection{Deformation and Magnetic Response}
\label{sec:transduction}

\begin{figure}[t]
\centering
\includegraphics[width=\columnwidth]{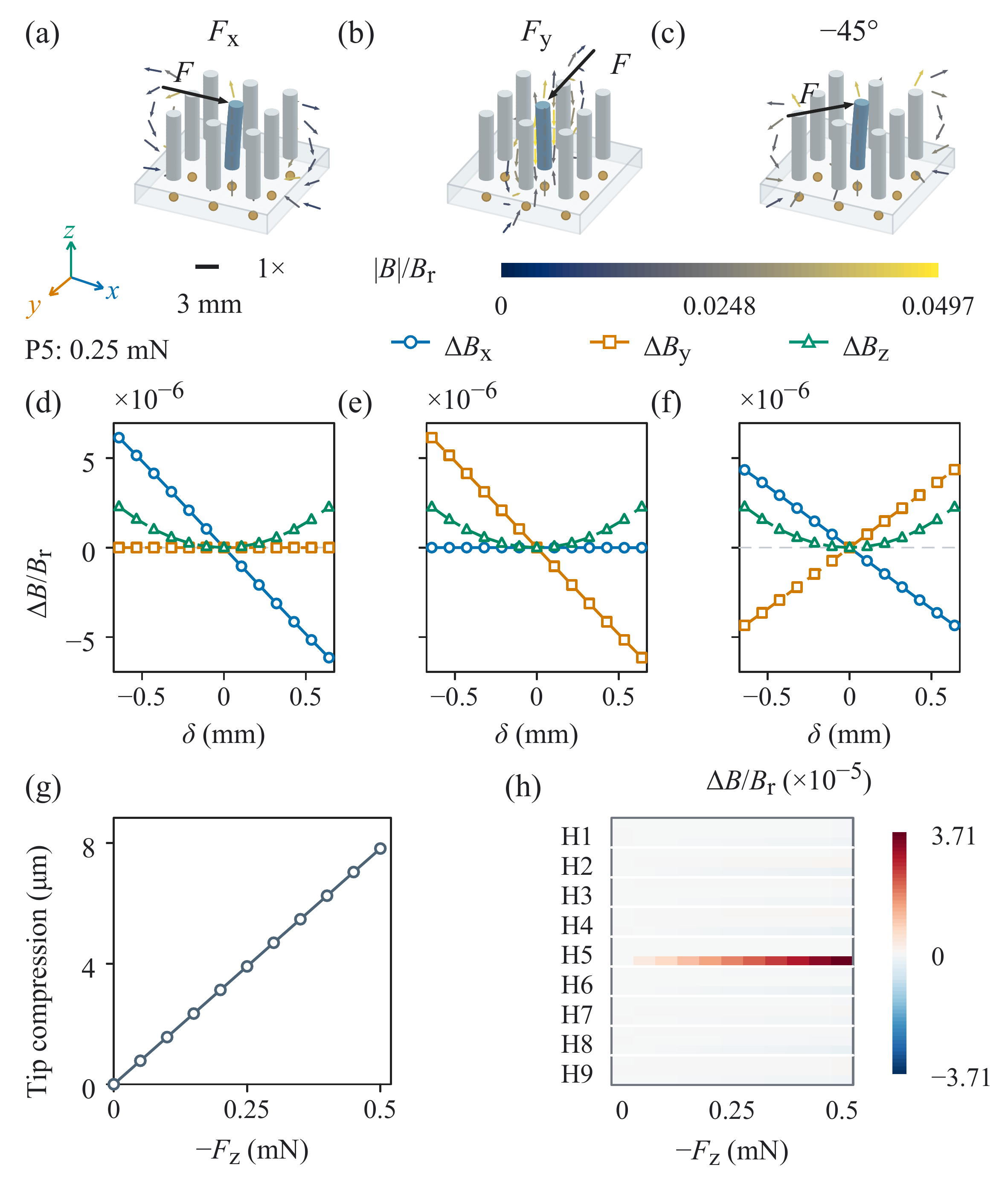}
\caption{Simulated deformation and magnetic response of the $3\times3$ array.
(a--c) True-scale geometries under 0.25~mN loads on P5 along $+x$, $+y$, and $-45^\circ$ in the $xy$ plane, with shared geometry and color scales. Arrow colors indicate $|B|/B_r$.
(d--f) H5 field increments versus projected mean tip displacement.
(g,h) P5 compression and all 27 field increments under normal loading, respectively; components in (h) are grouped by Hall point.}
\label{fig:sim_transduction}
\end{figure}

The center pillar P5 is subjected to lateral and normal loads to examine the effects of loading direction and deformation mode on the three-axis magnetic response at H5.
Lateral loads are applied along $x$, along $y$, and at $-45^\circ$ in the $xy$ plane (Fig.~\ref{fig:sim_transduction}(a--f)).
At 0.25~mN, the mean tip displacement projected along the loading direction is approximately 0.640~mm in all three cases, while the magnetic response varies with bending direction.
For bending along $x$ or $y$, the corresponding transverse field increment reverses sign when the displacement direction is reversed, whereas $\Delta B_z$ increases for bending in either direction.
Bending at $-45^\circ$ produces opposite signs in $\Delta B_x$ and $\Delta B_y$.
Under a normal compressive load of 0.5~mN, the mean tip compression is approximately $7.82\,\mu\mathrm{m}$, yielding $\Delta B_z/B_r\approx3.71\times10^{-5}$ at H5.
Figures~\ref{fig:sim_transduction}(g) and \ref{fig:sim_transduction}(h) show the load-dependent compression and all 27 magnetic-field increments, respectively.
These results demonstrate distinct magnetic responses to lateral bending and normal compression.

\subsection{Spatial Response and Two-Pillar Loading}
\label{sec:spatial}

\begin{figure}[t]
\centering
\includegraphics[width=\columnwidth]{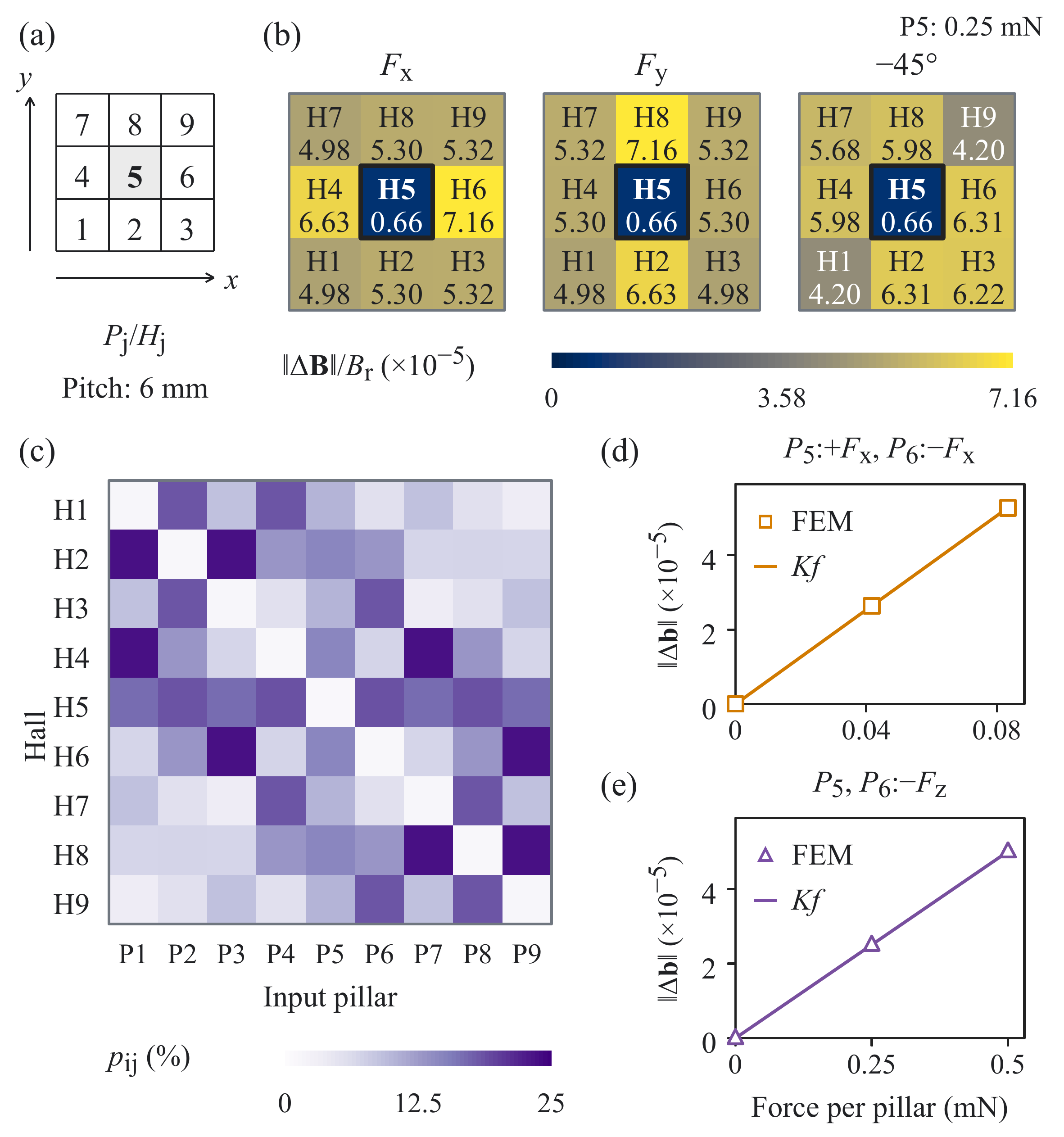}
\caption{Spatial response and two-pillar loading of the full array. (a) Pillar and Hall indices. (b) Field-increment norms at all nine Hall points under 0.25~mN loads on P5 along three lateral directions (shared color scale). (c) Normalized squared Frobenius norms of the response-matrix blocks. (d,e) Opposing $x$ shear and simultaneous normal compression of P5 and P6, respectively. Symbols denote nonlinear full-array outputs; lines denote local predictions $\mathbf K\mathbf f$. Both plots show the Euclidean norm of all 27 normalized field increments on a shared vertical scale.}
\label{fig:sim_array}
\end{figure}

Loading a single pillar induces magnetic-field changes at multiple Hall locations (Fig.~\ref{fig:sim_array}(a,b)).
For a 0.25~mN load along $+x$ on P5, the largest response occurs at the neighboring H6 rather than directly beneath the loaded pillar at H5, highlighting the spatially distributed nature of the magnetic response.

To characterize this distribution, small loads along $x$, $y$, and $z$ are applied separately to each pillar to estimate the local response matrix $\mathbf K$.
Each $3\times3$ block $\mathbf K_{ij}$ maps the force components on $P_j$ to the normalized field increments at $H_i$.
The relative squared sensitivity at each Hall location is defined as
\begin{equation}
p_{ij}
=\frac{\|\mathbf K_{ij}\|_F^2}
{\displaystyle\sum_{k=1}^{9}\|\mathbf K_{kj}\|_F^2},
\label{eq:2}
\end{equation}
where $\|\cdot\|_F$ denotes the Frobenius norm.
Figure~\ref{fig:sim_array}(c) presents $100p_{ij}$, showing how the sensitivity to each input pillar is distributed across the nine Hall locations.

Adjacent pillars P5 and P6 are then loaded simultaneously to compare the local linear prediction $\mathbf K\mathbf f$ with the nonlinear full-array response.
The relative difference over the complete 27-component response vector is defined as
\begin{equation}
e_{\mathrm{loc}}
=\frac{\left\|\Delta\widetilde{\mathbf b}_{\mathrm{FEM}}-\mathbf K\mathbf f\right\|_2}
{\left\|\Delta\widetilde{\mathbf b}_{\mathrm{FEM}}\right\|_2},
\label{eq:3}
\end{equation}
where $\Delta\widetilde{\mathbf b}_{\mathrm{FEM}}$ is the nonlinear response and $\|\cdot\|_2$ denotes the Euclidean norm.
For the tested loads, the difference remains below 1.7\% under opposing lateral loading and 0.12\% under normal compression (Fig.~\ref{fig:sim_array}(d,e)).

These results show that local pillar loading produces a distributed magnetic response that can be characterized using the full Hall array.
The local linear model closely approximates the nonlinear response within the tested load range, supporting its use for describing small-load interactions across the array.

\section{Robotic Surface Perception Experiments}
\label{sec:experiments}

\begin{figure*}[t]
\centering
\includegraphics[width=0.96\textwidth]{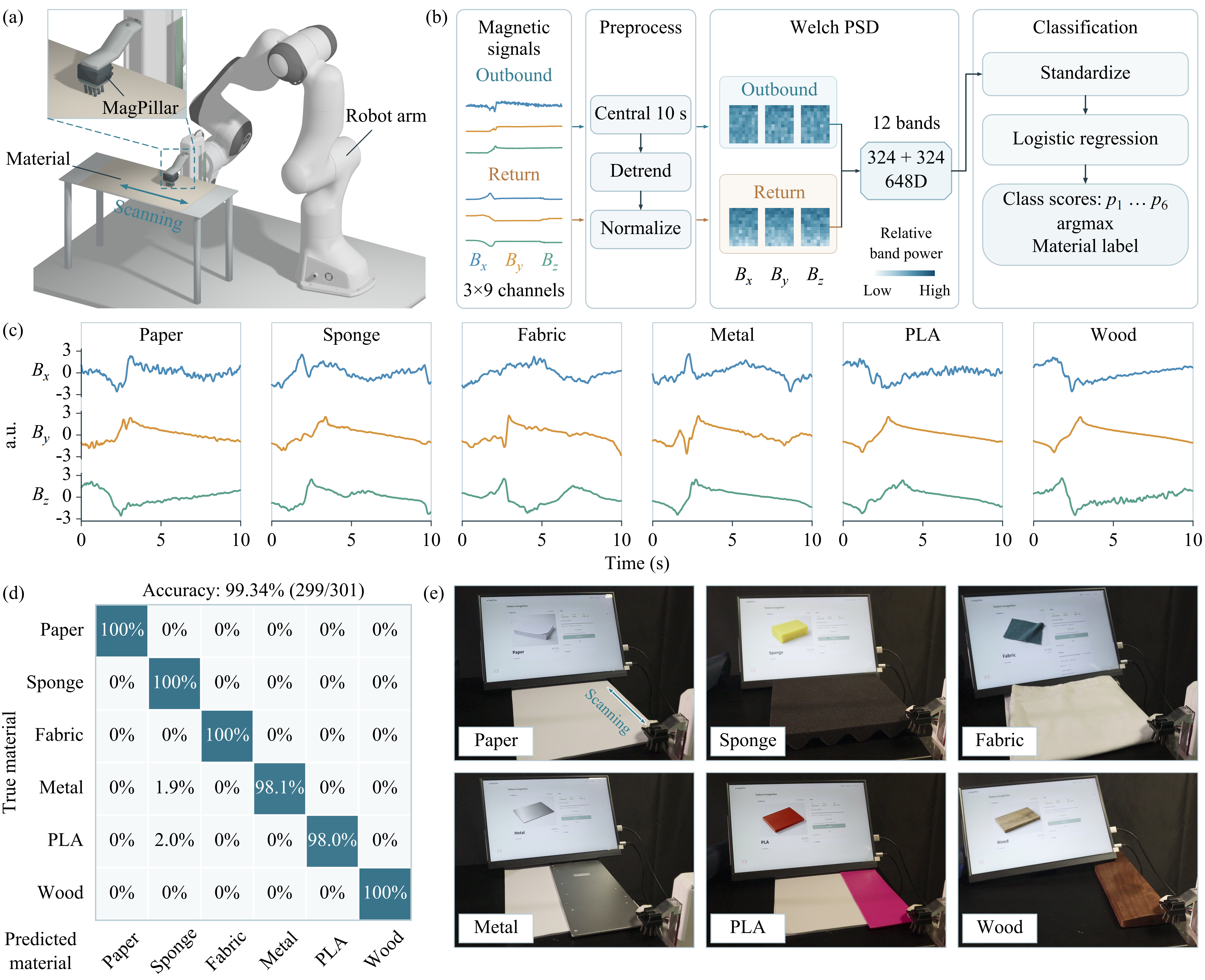}
\caption{Robotic material surface recognition using MagPillar. (a) Scanning setup and contact. (b) Spectral-feature and logistic-regression pipeline. (c) Representative rank-normalized m5 signals from the central 10~s of outbound scans. (d) Row-normalized confusion matrix for within-session five-fold temporally blocked cross-validation: 99.34\% (299/301). (e) Online recognition examples for six materials.}
\label{fig:demo1}
\end{figure*}

\begin{figure*}[t]
\centering
\includegraphics[width=0.975\textwidth]{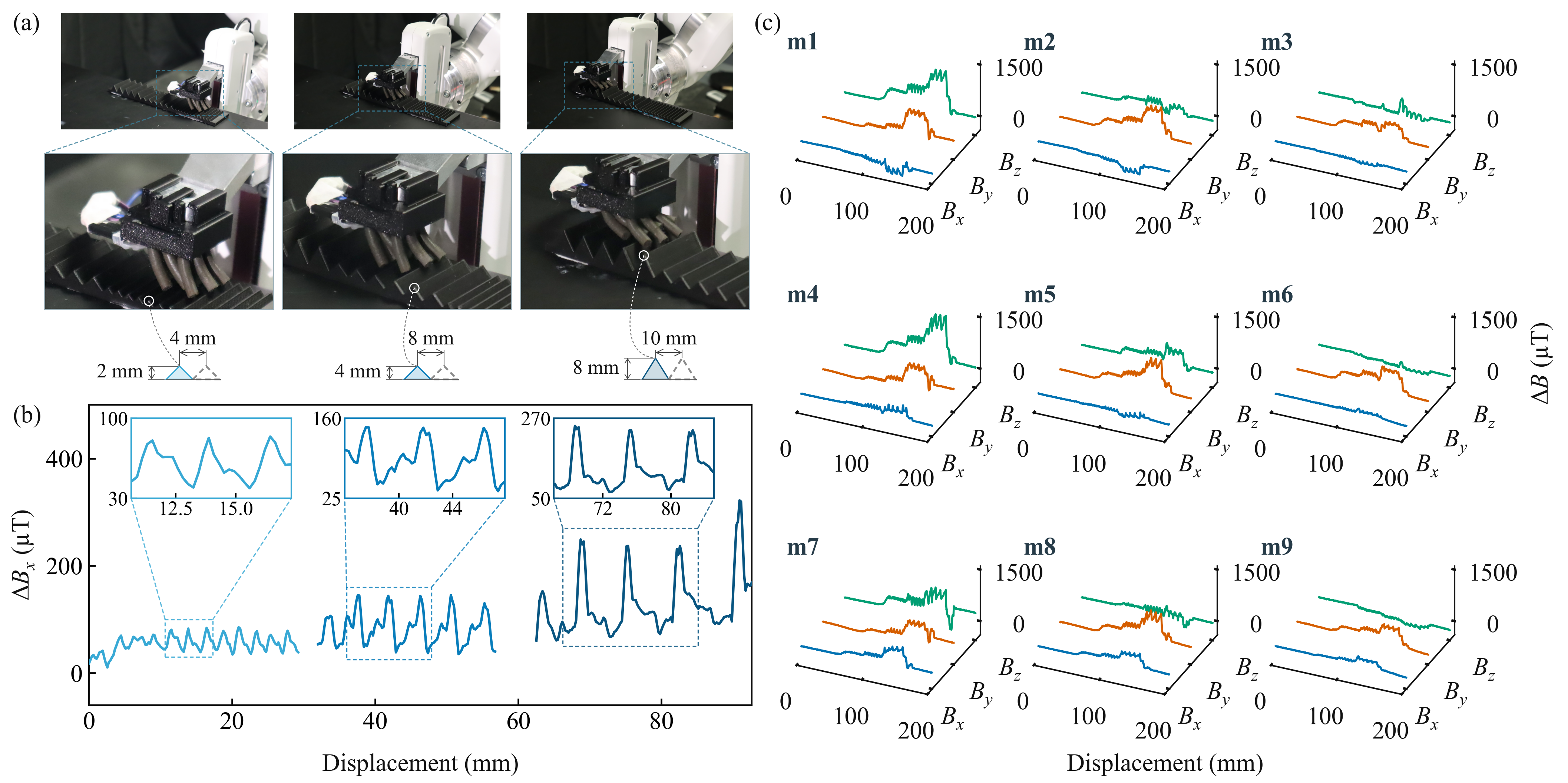}
\caption{Continuous sensing of multiscale protrusions.
(a) Experimental photographs and profile dimensions.
(b) Representative $\Delta B_x$ responses of m4 with enlarged views.
(c) Tri-axial responses of the $3\times3$ array.}
\label{fig:demo2}
\end{figure*}

To evaluate MagPillar during robotic scanning, the sensor was mounted on the gripper of a Franka Research 3 robot arm and scanned across test specimens with its pillars in contact with their surfaces, as shown in Figs.~\ref{fig:demo1}(a) and~\ref{fig:demo2}(a). All 27 magnetic-field signals were acquired at approximately 30~Hz and recorded by a host computer. The experiments comprised material recognition using round-trip scans and surface-profile sensing using continuous scans over periodic protrusions.

\subsection{Material Surface Recognition}
\label{sec:materials}
Material recognition was evaluated using six surfaces: A4 paper, sponge, fabric, steel (labeled ``Metal'' in Fig.~\ref{fig:demo1}), polylactic acid (PLA), and wood. Differences in mechanical compliance, surface roughness, and texture can influence the contact responses recorded during sliding. Approximately 50 complete round-trip scans were collected for each material, with each round trip treated as one classification sample.

The recognition pipeline is shown in Fig.~\ref{fig:demo1}(b). For each round trip, the central 10~s of the outbound and return scans were extracted separately and linearly resampled to 30~Hz, yielding 300 time points per direction. Within each segment, the channels underwent linear detrending, a rank-to-Gaussian transformation, and a second linear detrending step. Representative normalized three-axis signals (a.u.) from sensing unit m5 exhibit distinct fluctuation patterns across the six tested surfaces during outbound scans (Fig.~\ref{fig:demo1}(c)). For each channel, the power spectral density was estimated using Welch's method. Relative powers in 12 frequency bands per channel were log-transformed and combined across all 27 channels and both scan directions to form a 648-dimensional feature vector per round trip. The feature vectors were standardized using parameters fitted only on the training data and passed to a logistic regression classifier implemented in scikit-learn.

For evaluation, the records from each data collection session were divided into five consecutive groups in acquisition order. In each of five rounds, the group with the same index from each session was used for testing, while the remaining records were used for training. The round trips immediately before and after each test group were excluded from training. Each sample was tested once, and predictions from all five rounds were pooled. The overall offline classification accuracy reached 99.34\% (299/301), with predictions concentrated along the main diagonal of the confusion matrix in Fig.~\ref{fig:demo1}(d), indicating effective discrimination among the six tested material surfaces.

Fig.~\ref{fig:demo1}(e) further presents online recognition demonstrations for all six materials. In these demonstrations, magnetic responses acquired during robotic scanning were processed by the recognition pipeline, and the predicted material labels were displayed on the host interface, illustrating the integration of tactile data acquisition and material classification. Together, these results demonstrate MagPillar's ability to convert material-dependent contact interactions into discriminative magnetic signals, supporting tactile material recognition during robotic surface exploration.

\subsection{Continuous Contact Sensing over Protrusions}
\label{sec:protrusions}

The compliant magnetic pillars deform as they slide over surface features, converting variations in surface geometry into measurable magnetic-field changes. To evaluate the sensor's response to different surface profiles, a specimen was designed with three groups of periodic protrusions having nominal heights of 2, 4, and 8~mm and corresponding pitches of 4, 8, and 10~mm, respectively. The robot arm scanned the sensor across the specimen, as shown in Fig.~\ref{fig:demo2}(a).

Fig.~\ref{fig:demo2}(b) presents the representative $\Delta B_x$ response of sensing unit m4, with repeated peaks arising from successive contacts with the protrusions. The fluctuation amplitude and spacing between adjacent major peaks generally increase across the three regions, consistent with the increasing height and pitch of the surface features. This correspondence indicates that the magnetic signals capture variations in surface geometry and provide tactile cues for inferring the local surface profile. Fig.~\ref{fig:demo2}(c) further shows the three-axis magnetic responses of all nine sensing units during the same scan. Their distinct amplitudes and waveforms reveal the spatial distribution of the magnetic response to contact across the array, providing complementary information for continuous perception of surface contours and geometric features.

\section{Conclusion}
\label{sec:conclusion}

This work presented MagPillar, a tactile sensor that combines a passive $3\times3$ compliant magnetic pillar array with nine three-axis Hall sensors for robotic surface perception. Mechanical and magnetic-field analysis of the complete array related pillar deformation to the spatial responses of all 27 magnetic-field channels. The simulations showed that the strongest response can occur at a Hall location adjacent to the loaded pillar. Robotic scanning experiments achieved an offline accuracy of 99.34\% in distinguishing six material surfaces under within-session temporally blocked cross-validation. Continuous scanning over periodic protrusions with different heights and pitches produced corresponding variations in signal amplitude and peak spacing. These results demonstrate the use of compliant magnetic pillars and distributed Hall sensing for material recognition and continuous sensing of surface features. Future work will explore broader applications of MagPillar, including water-flow and airflow sensing, wearable monitoring of human motion, and integration with robotic grippers for closed-loop surface exploration and contact-guided manipulation.

\let\originalthebibliography\thebibliography
\renewcommand{\thebibliography}[1]{\originalthebibliography{#1}\interlinepenalty=10000}
\bibliographystyle{IEEEtran}
\bibliography{references}
\end{document}